\documentclass[10pt]{article}

\usepackage[utf8]{inputenc} 
\usepackage[T1]{fontenc}    
\usepackage{times}          
\usepackage{url}
\usepackage[colorlinks=true,linkcolor=blue,citecolor=blue,urlcolor=blue]{hyperref} 
\usepackage{comment}
\usepackage{xpatch}
\usepackage[toc,page]{appendix}
\usepackage{graphicx}%
\usepackage{amsmath,amssymb,amsfonts}%
\usepackage{amsthm}%
\usepackage{mathrsfs}%
\usepackage{xcolor}%
\usepackage{textcomp}%
\usepackage{manyfoot}%
\usepackage{booktabs}%
\usepackage{algorithm}%
\usepackage{algorithmicx}%
\usepackage{algpseudocode}%
\usepackage{listings}%
\usepackage{colortbl}
\usepackage{todonotes}
\usepackage{makecell}
\usepackage{multirow}
\usepackage{orcidlink}
\usepackage{siunitx}
\usepackage{subcaption}
\usepackage{multirow, makecell}

\title{Machine Unlearning for Speaker-Agnostic Detection of Gender-Based Violence Condition in Speech}

\author{%
    Emma Reyner-Fuentes\orcidlink{0009-0004-4307-0057}\thanks{Corresponding author: emma.reyner@uc3m.es. \newline Department of Signal Theory and Communications, University Carlos III of Madrid (UC3M), ROR: \url{https://ror.org/03ths8210}, Leganés, Madrid, Spain.}
    \and
    Esther Rituerto-González\orcidlink{0000-0001-5597-4556}\thanks{Department of Psychiatry and Psychotherapy, University Hospital, Ludwig-Maximilian-University of Munich, Germany, \newline German Center for Mental Health (DZPG), partner site Munich-Augsburg, Germany}
    \and
    Carmen Peláez-Moreno\orcidlink{0000-0003-1425-6763}\thanks{Department of Signal Theory and Communications, University Carlos III of Madrid (UC3M), ROR: \url{https://ror.org/03ths8210}, Leganés, Madrid, Spain, \newline Institute of Gender Studies, Universidad Carlos III de Madrid (UC3M), ROR: \url{https://ror.org/03ths8210}, Leganés, Madrid, Spain}
}

\date{} 

\begin{document}

\maketitle

\thispagestyle{empty}


\begin{abstract}
Gender-based violence is a pervasive public health issue that severely impacts women's mental health, often leading to conditions such as in anxiety, depression, post-traumatic stress disorder, and substance abuse. Identifying the combination of these various mental health conditions could then point to someone who is a victim of gender-based violence. And while speech-based artificial intelligence tools show as a promising solution for mental health screening, their performance often deteriorates when encountering speech from previously unseen speakers, a sign that speaker traits may be confounding factors.
This study introduces a speaker-agnostic approach to detecting the gender-based violence victim condition from speech, aiming to develop robust artificial intelligence models capable of generalizing across speakers. By employing domain-adversarial training, we reduce the influence of speaker identity on model predictions, we achieve a 26.95\% relative reduction in speaker identification accuracy while improving gender-based violence victim condition classification accuracy by 6.37\% (relative). These results suggest that our models effectively capture paralinguistic biomarkers linked to the gender-based violence victim condition, rather than speaker-specific traits.
Additionally, the model’s predictions show moderate correlation with pre-clinical post-traumatic stress disorder symptoms, supporting the relevance of speech as a non-invasive tool for mental health monitoring. This work lays the foundation for ethical, privacy-preserving artificial intelligence systems to support clinical screening of gender-based violence survivors.
\end{abstract}


\section{Introduction}

Gender-based violence (GBV)--encompassing physical, sexual, and psychological harm directed at individuals due to their gender--represents a widespread social and public health problem. Women and girls are disproportionately affected by GBV, with long-lasting consequences for their physical and mental well-being~\cite{eige2023}.
Research studies consistently link GBV with adverse psychological outcomes, including anxiety, depression, suicidal ideation, substance use, and, in particular, post-traumatic stress disorder (PTSD) \cite{oram2017violence, escriba2010IPV, ferrari2014DVA, spencerHealthEffectsAssociated2023, haeringDisentanglingSexDifferences2024}. PTSD is often the most prevalent mental health condition among GBV survivors \cite{ferrari2014DVA, chandan2019ipv, shen2019intimate}, with symptoms that impair social functioning, daily activities, and overall quality of life.

According to the World Health Organization (WHO) and the United Nations (UN), a GBV victim is a woman subjected to violence that causes or risks causing physical, sexual, or psychological harm, including threats, coercion, or arbitrary restrictions of liberty, regardless of the context~\cite{who2021, undoc1993}.
Some national legislations, such as Spain's Organic Law 1/2004~\cite{leyorganica2004}, further recognize the indirect impact on children who witness such violence as victims of "vicarious violence". 

Psychologists and social services' professionals typically identify GBV victimization through clinical interviews or self-reported questionnaires, such as the EGS-R for PTSD assessment~\cite{echeburua2016egsr}. However, this process requires victims to disclose past abuse early, an obstacle given the stigma, fear, or denial that often surrounds these experiences. Underreporting remains a serious challenge, leaving many cases unidentified \cite{shanmugamQuantifyingDisparitiesIntimate2024, spencerHealthEffectsAssociated2023, chenUsingMachineLearning2023}.

In this context, speech-based artificial intelligence (AI) tools offer a promising, non-invasive approach to support mental health screening. These tools can be integrated into virtual assistants, therapy applications, or helplines to unobtrusively assess emotional and mental health states. Speech has already proven useful for diagnosing depression~\cite{koops2023speech}, suicide risk~\cite{scherer2013suicidal, belouali2021acoustic}, and other physical and mental conditions~\cite{latif2020speech, wanderley2022detection}. Unlike traditional assessments, speech-based systems may reduce bias by not relying on explicit self-report, thus improving accessibility and early detection.

Machine Learning (ML) and, more specifically, Deep Learning (DL) models can extract meaningful paralinguistic features from speech signals, such as tone, pitch, rhythm, and vocal variability, beyond linguistic content~\cite{schuller2013paralinguistics}. However, these models often encode speaker-specific traits, which can degrade models' generalization and raise privacy concerns in clinical contexts. Although personalization may improve performance in tasks such as speech emotion recognition (SER)~\cite{mura2020ser}, it may introduce undesirable confounds and ethical issues in mental health applications.

Building on our previous work~\cite{reyner22iberspeech, reyner23interspeech}, we hypothesize that speaker identity contributes undesirably to GBVVC detection models. Therefore, in order to attain a speaker-agnostic detector that can distinguish between GBV victims and non-victims, we explore the use of domain-adversarial training techniques to disentangle relevant from irrelevant speaker-related information.

Domain-Adversarial Neural Networks (DANNs)~\cite{ganin2016domainadversarial} are used as a novel approach for domain adaptation by learning to ignore domain-sensitive characteristics that are inherently embedded in the data.
These networks aim to actively learn feature representations that exclude information on the data domain.
This methodology promotes the development of features that are discriminative for a main task while remaining domain-agnostic.

\begin{figure}[h]
\centering
\includegraphics[width=0.6\textwidth]{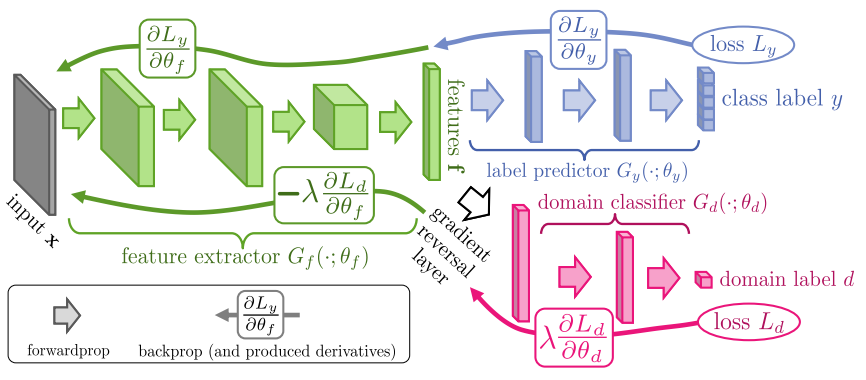}
\caption{Original Domain-Adversarial Neural Network architecture~\cite{ganin2016domainadversarial}.}
\label{fig:dann_orig}
\end{figure}

DANNs, as defined in~\cite{ganin2016domainadversarial}, consist of two classifiers: one for the \textit{main} classification task and another for \textit{domain} classification. Figure~\ref{fig:dann_orig} represents the original DANN architecture scheme. These classifiers share initial layers that shape the data representation. A gradient reversal layer is introduced between the domain classifier and the feature representation layers so that it forwards the data during forward propagation and reverses the gradient signs during backward propagation. In this manner, the network aims to minimize classification error for the main task while maximizing the error of the domain classifier, ensuring an effective and discriminative representation for the primary task, in which domain information is suppressed. 

In our specific context, the main or primary task entails GBVVC classification, while the domain classification  refers to speaker identity. We aim to ensure that the feature representation remains discriminative for the primary task while eliminating any traces of speaker-identity information.

Speaker-agnostic models offer several advantages: they prevent the identification of individual speakers, thus preserving privacy in mental health applications. Additionally, these models are designed to be generalizable and independent of the speaker's identity, making them effective for an extensive range of users, even those who have never been seen by the training phase of the models. By intentionally not learning speaker-specific information, they focus on improving performance for the task at hand, ensuring that the model's resources are dedicated only to learning relevant information.

We evaluate our method on an extended version of the WEMAC database~\cite{miranda2024wemac}, which includes spontaneous speech samples and emotional self-assessments from women in response to audiovisual stimuli. The extended dataset comprises additional speech data from 39 GBV survivors, whose participation was ethically approved but whose data remains unpublished due to privacy concerns. All participants were screened with the EGS-R questionnaire, and only survivors with pre-clinical PTSD symptoms (scores $\leq 20$) were included, ensuring the sample represented recovered individuals. Despite this, our system was still able to detect subtle residual signals associated with past victimization. More details on the dataset and feature extraction methodology are presented in Section~\ref{sec:database}.

Our findings suggest that domain-adversarial learning effectively suppresses speaker-specific cues and enhances the models' ability to focus on paralinguistic indicators of GBVVC. By promoting generalizability and privacy, this work contributes to the development of trustworthy AI tools for sensitive mental health applications.

\section{Results}

\subsection*{Prior Work}

We previously explored the detection of gender-based violence victim condition (GBVVC) from speech signals in our work \cite{reyner22iberspeech, reyner23interspeech}. In the first paper \cite{reyner22iberspeech}, we demonstrated that using paralinguistic features extracted from speech, it is possible to differentiate between victims and non-victims with an accuracy of 71.53\% under a speaker-independent (SI) setting. In our subsequent work \cite{reyner23interspeech}, we incorporated additional data and investigated the relationship between model performance and self-reported psychological and physical symptoms. The system achieved a user-level SI accuracy of 73.08\% via majority voting over individual utterances. However, in both studies, although we employed a Leave-One-Subject-Out (LOSO) strategy, our hypothesis is that speaker-related traits remained entangled with the target GBVVC labels. We hypothesized that based on the consistently lower performance observed under the LOSO evaluation scheme, compared to other data-splitting strategies, which we believe indicated that the model was inadvertently leveraging speaker-specific characteristics to perform the classification task. This potential confound motivated the present work, where we aim to remove speaker-specific information while preserving task-relevant patterns. As we will further discuss in in section \ref{sec:discuss}, several authors have already warned about the prevalence of this kind of problem in speech related research for health diagnosis where data is less abundant than for other speech technologies.

\subsection{Baseline Models}
Since this is, to the best knowledge of the authors, the first study addressing this specific task, we established a set of baseline models to set a reference performance and, therefore, contextualize the performance of the proposed Domain-Adversarial Model (DAM). Although our previous models \cite{reyner22iberspeech, reyner23interspeech} focused on GBVV classification, they were not designed with an adversarial structure to explicitly enforce speaker invariant feature learning. Hence, they are not directly comparable to the DAM approach. The baseline models introduced here serve this role by providing a clear benchmark under similar experimental conditions.

To this end, we first trained two baseline models using speech embeddings: (1) the Isolated Condition Model (ICM), which predicts GBVVC status, and (2) the Isolated Speaker Model (ISM), which identifies speaker identity. These models--whose architecture is depicted in Figure \ref{fig:figiso}
--are trained on frame-level speech segments of 1~s duration. Majority voting (MV) is then applied at the user level to obtain final predictions per subject.

\begin{figure}[htpb]
\centering
\begin{subfigure}[b]{0.32\textwidth}
\centering
\includegraphics[width=\textwidth]{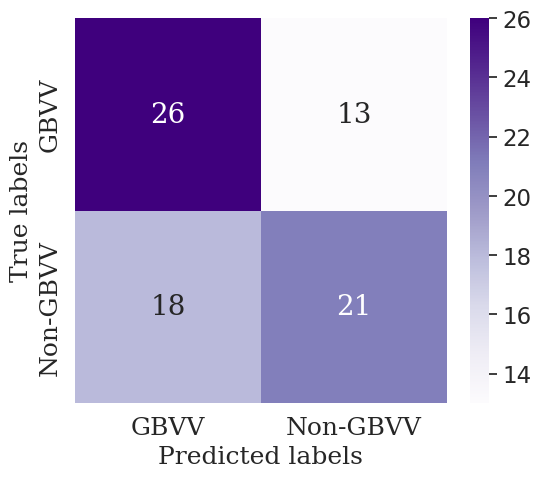}
\caption{User-level confusion matrix}
\label{fig:CMuserICM}
\end{subfigure}
\hspace{0.05\textwidth}
\begin{subfigure}[b]{0.35\textwidth}
\centering
\includegraphics[width=\textwidth]{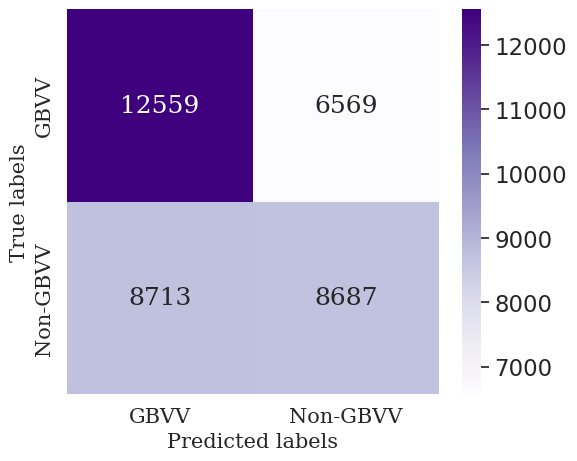}
\caption{Frame-level confusion matrix}
\label{fig:CMframeICM}
\end{subfigure}
\caption{Confusion matrices for the Isolated Condition Model (ICM).}
\label{fig:CM_ICM}
\vspace{-1em}
\end{figure}

The ICM model achieves a frame-level accuracy of $58.53\%$, with a precision of $59.04\%$, recall of $59.28\%$, and F1-score of $59.16\%$. At the user-level, the ICM reaches an accuracy of $60.00\%$, with a precision of $61.90\%$, recall of $59.09\%$, and F1-score of $60.47\%$. These results are presented in the corresponding confusion matrices, in Figure~\ref{fig:CM_ICM}. The figure presents both user-level and frame-level confusion matrices for a visual assessment of model behavior across scales.

In contrast, the ISM model, designed to identify speaker identity from speech, achieves a frame-level accuracy of $91.34\%$ (Table~\ref{tab:results}), underscoring the strong presence of speaker-specific information in the embeddings. However, since this model performs a multi-class classification task involving 78 different speakers, a confusion matrix or performance metrics such as precision, recall, or F1-score at the user level are not meaningful or directly comparable. Each prediction corresponds to a specific speaker identity rather than a binary classification problem, thus precluding the application of majority voting or binary confusion-based metrics.

\subsection{Speaker-Agnostic GBVVC Detection}

In this section, we introduce our Domain-Adversarial Model (DAM), specifically designed to remove speaker-dependent features from the learnt speech embeddings, thereby promoting the emergence of representations that are informative for GBVVC detection but invariant to speaker identity. This is achieved through a gradient reversal layer that enables adversarial training between the main GBVVC classification task and an auxiliary speaker-identification task.

\begin{figure}[htpb]
\centering
\begin{subfigure}[b]{0.35\textwidth}
\centering
\includegraphics[width=\textwidth]{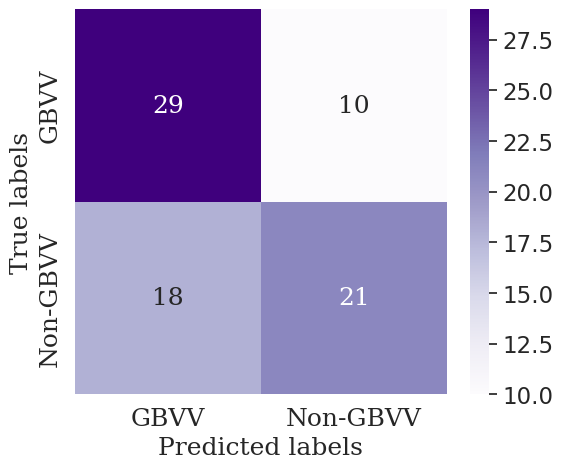}
\caption{User-level}
\label{fig:CMiuserDAM}
\end{subfigure}
\hspace{0.05\textwidth}
\begin{subfigure}[b]{0.35\textwidth}
\centering
\includegraphics[width=\textwidth]{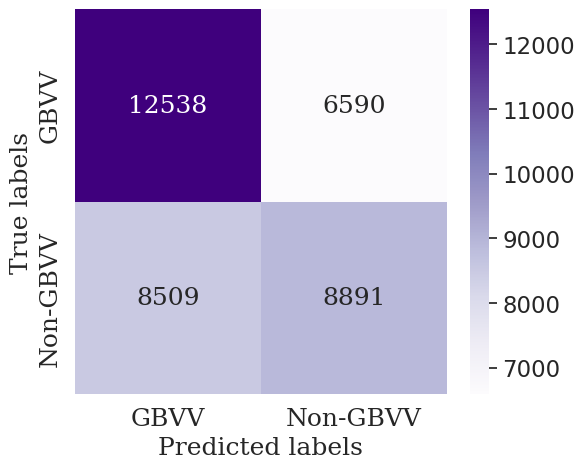}
\caption{Frame-level}
\label{fig:CMaframeDAM}
\end{subfigure}
\caption{Confusion matrices for the Domain-Adversarial Model (DAM).}
\label{fig:CM_DAM}
\vspace{-1em}
\end{figure}

Figure~\ref{fig:CM_DAM} presents the confusion matrices at both user- and frame-levels for the main task of the DAM. At the user-level, DAM achieves an accuracy of $64.10\%$, precision of $61.70\%$, recall of $74.36\%$, and F1-score of $67.44\%$. At the frame-level, the model yields an accuracy of $58.66\%$, precision of $59.57\%$, recall of $65.55\%$, and F1-score of $62.42\%$. These results reveal a more balanced classification performance compared to the ICM.
Specifically, DAM improves upon the ICM by a relative $+1.14\%$ in frame-level accuracy (from $58.11\%$ to $58.66\%$) and $+6.37\%$ in user-level accuracy (from $60.26\%$ to $64.10\%$). It also outperforms the baseline in F1-score at both levels: $+0.99$ points at frame-level and $+4.06$ points at user-level, indicating improved balance between precision and recall.

Nonetheless, the dataset is balanced perfectly at subject level and quite evenly at frame level ($52.36\%$ GBVVC vs. $47.64\%$ non-GBVVC), which validates the use of accuracy as the primary evaluation metric. Results are reported in Table~\ref{tab:results}, with mean and standard deviation of the accuracy across 78 LOSO folds, reflecting the real-world variability due to differences in recording quality, vocal traits, and symptom expression. High inter-subject variance is expected in this kind of clinical speech dataset and the fact that our DAM approach reduces the standard deviation (STD) in the GBVVC classification task while increasing the STD in the speaker classification task highlights its effectiveness in promoting speaker-invariant representations and enhancing the model’s generalization capability.

Table~\ref{tab:results} compares four models across two tasks: GBVVC detection and speaker identification. The first column shows the performance of the ICM, which acts as our non-adversarial baseline. While we previously reported better results \cite{reyner23interspeech}, those models are not directly compatible with the ones here, trained in an adversarial learning fashion. Hence, the ICM serves as a fairer baseline for this analysis.
The second column shows the results of the DAM, which, despite the adversarial unlearning of speaker identity, maintains and slightly improves GBVVC classification performance. This suggests that speaker-related information is not essential--or potentially even detrimental--for detecting the condition.

\begin{figure}[htpb]
\centering
\includegraphics[trim={0cm 0cm 10cm 14.7cm},clip,width=0.75\textwidth]{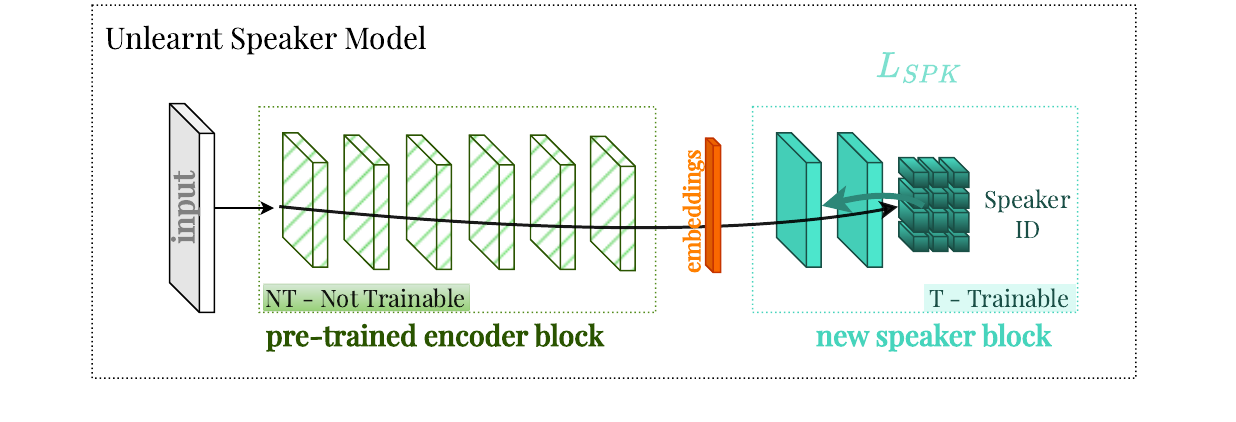}
\caption{Unlearnt Speaker Model Architecture. The black, thinner arrows correspond to the forward step. The thicker, colored arrows correspond to the backpropagation steps.}
\label{fig:fig_new_spk}
\end{figure}

The third and fourth columns assess the extent to which speaker identity is removed from the learnt representations. The Isolated Speaker Model (ISM), trained directly to identify speakers, achieves $91.34\%$ accuracy, confirming that speaker traits are easily learnable from feature embeddings. The fourth column shows the results for the the Unlearnt Speaker Model (USM), which uses the same architecture but operates on the frozen embeddings from the DAM, (see Figure~\ref{fig:fig_new_spk}). The USM however, shows a steep drop in speaker identification accuracy to $66.72\%$, a relative degradation of $26.95\%$. This demonstrates the success of the adversarial strategy in attenuating speaker-specific information from the learnt feature embeddings.

We emphasize that speaker identification models cannot be evaluated under a true LOSO scheme because, by definition, they cannot classify previously unseen identities. Therefore, we use a subject-dependent evaluation strategy with less metrics for ISM and USM, ensuring that no segments from the same audio recording are shared between training and test partitions. This avoids data leakage while preserving speaker identity in the training data.

The attenuation of speaker information by over $26\%$ does not harm GBVVC detection. On the contrary, it enhances generalization, yielding a notable $6.37\%$ improvement in subject-level MV accuracy. This strongly supports our hypothesis that speaker traits act as a confounding variable, and their removal allows the model to better attend to condition-relevant vocal biomarkers. We believe that refining our adversarial strategy to achieve even greater disentanglement of speaker identity could further amplify these gains in future iterations.

\begin{table}[h!]
\renewcommand{\arraystretch}{1.15}
\centering
\resizebox{0.8\columnwidth}{!}{
\begin{tabular}{c || c | c || c | c || c | c}
\hline
 & & \multicolumn{2}{c||}{GBVVC Detection} & \multicolumn{2}{c}{Speaker Identification} \\
\hline
 & & ICM & DAM & ISM & USM \\
\hline \hline
\multirow{5}{*}{\rotatebox{90}{\centering FL}} 
& Mean Accuracy       & $58.11 \pm 31.85$ & $58.66 \pm 31.52$ & $91.34 \pm 3.72$ & $66.72 \pm 6.71$ \\
& Precision           & $58.18$ & $59.57$ & -     & -     \\
& Recall              & $64.87$ & $65.55$ & -     & -     \\
& F1-Score            & $61.43$ & $62.42$ & -     & -     \\
\hline
\multirow{2}{*}{\rotatebox{90}{\centering UL}} 
& Accuracy            & $60.26$ & $64.10$ & -     & -     \\
& F1-Score            & $63.38$ & $67.44$ & -     & -     \\
\hline
\end{tabular}
}
\caption{Comparison of frame (FL)- and user (UL) -level metrics across models. ICM: Isolated Condition Model. DAM: Domain-Adversarial Model. ISM: Isolated Speaker Model. USM: Unlearnt Speaker Model.}
\label{tab:results}
\end{table}

\subsection{The correlation with EGS-R score}
To further understand what cues the model is leveraging to make GBVVC predictions, we explored the relationship between model performance and clinical symptomatology, as measured by the EGS-R score. This scale, ranging from 0 to 20 in our case, quantifies pre-clinical PTSD-related symptoms such as re-experiencing, emotional numbing, and hyperarousal.

In our previous work \cite{reyner23interspeech}, we found that correctly classified GBVVC subjects tended to have higher EGS-R scores, suggesting that our models implicitly relied on speech markers associated with trauma-related symptomatology. Here, we replicate this analysis for both the ICM and DAM models by comparing the mean EGS-R scores of correctly and incorrectly classified GBVVC subjects.

\begin{table}[h!]
\renewcommand{\arraystretch}{1.15}
\centering
\resizebox{.8\columnwidth}{!}{
\begin{tabular}{c | c | c | c | c}
 & \multicolumn{2}{c}{Domain-Adversarial Model (DAM)} & \multicolumn{2}{|c}{Isolated Condition Model (ICM)} \\
\hline
 & Correctly Classified & Misclassified & Correctly Classified & Misclassified \\
\hline
Mean EGS-R & $10.52$ & $6.90$ & $10.35$ & $8.07$ \\
Scaled (\%) & $52.59$ & $34.50$ & $51.73$ & $40.39$ \\
\hline
\end{tabular}}
\caption{Mean EGS-R score for GBVVC subjects by classification outcome and model. EGS-R is only available for GBVVC participants.}
\label{tab:egsr_results}
\end{table}

Table~\ref{tab:egsr_results} shows a statistically significant gap in symptom severity between correctly and incorrectly classified victims for both models (DAM: $p = 0.0017$, ICM: $p = 0.0351$, t-Student based), this is, that the mean EGS-R score for correctly classified subjects is significantly higher than that of the subjects who have been incorrectly classified. This reinforces the notion that our models are not making arbitrary predictions, but instead exploit subtle voice markers correlated with PTSD-like symptoms. Furthermore, the difference in EGS-R scores between correct and incorrect classifications increases under the DAM, although not significantly ($p = 0.0711$, t-Student based), indicating that the adversarial training may have the potential to enhance the model’s focus on trauma-related vocal features once speaker confounds are removed.

Taken together, these findings align with our broader hypothesis: the ability to detect GBVVC from speech is closely tied to the manifestation of early trauma symptoms, which can be encoded in paralinguistic cues such as pitch variability, jitter, or prosodic flattening. By eliminating speaker-dependent information, the DAM enables a clearer identification of these subtle cues, contributing to both improved accuracy and better interpretability grounded in clinical theory.

\section{Discussion}
\label{sec:discuss}

The automatic detection of gender-based violence victim condition (GBVVC) through speech analysis remains a nascent and underexplored research area. Most prior work in this field has focused on detecting GBV from textual data, particularly through the analysis of social media posts using natural language processing (NLP) techniques \cite{math9080807, 8685083, yallico2022gbv}. To the best of our knowledge, no previous studies have attempted to identify the condition of GBV victims directly from their speech, apart from our prior contributions \cite{reyner22iberspeech, reyner23interspeech}.

In our earlier work, we demonstrated that machine learning (ML) models can distinguish between speech samples from GBV victims and non-victims. However, we identified a major confounding factor: the models tended to rely on speaker identity cues, performing speaker identification alongside the detection of speech traits associated with the GBVVC. This phenomenon is widely acknowledged in ML-based clinical diagnostics, especially in speech-related applications, where datasets are typically small and prone to overfitting to speaker-specific features \cite{berishaAreReportedAccuracies2022, rutowskiCorpusSizeRequirements2022}. As highlighted in \cite{millingSpeechNewBlood2022}, such biases present a significant barrier to the clinical applicability of ML technologies in mental health and diagnostic settings.
To address this issue, we pursued a speaker-agnostic approach that mitigates the influence of speaker-specific information in the classification task. This direction aligns with emerging trends in ML such as "Machine Unlearning", which aims to erase irrelevant or potentially biased information from model representations \cite{xuMachineUnlearningSurvey2023}. In our case, the objective was to remove speaker identity from the learnt embeddings, thereby ensuring that predictions of GBVVC are not driven by user-specific traits and enhancing model generalizability and privacy.

Inspired by domain-adversarial training methods previously used to disentangle speaker traits from emotion in speech emotion recognition (SER) tasks \cite{li2020adversarial, han2018adversarial}, we adapted the DANN (Domain-Adversarial Neural Network) framework to our task. By treating each speaker as a domain, we trained an encoder to learn features that are discriminative for GBVVC classification while being invariant to speaker identity. This was accomplished via adversarial training using a gradient reversal layer that discourages the encoder from encoding speaker-related information.

The implementation of this model achieved two main outcomes. First, it reduced the capacity of the model to identify speakers by 26.95\%, validating that speaker-specific information had been effectively suppressed. Second, this reduction coincided with a relative 6.37\% improvement in GBVVC classification accuracy, as shown in Table~\ref{tab:results}. These results support our hypothesis that removing speaker-specific information allows the model to focus on more relevant, potentially clinical, acoustic patterns associated with the GBVVC. In doing so, we alleviate previous concerns about the model’s over-reliance on speaker identity and advance towards robust, speaker-independent speech biomarkers. 

Additionally, the analysis of EGS-R scores (Table~\ref{tab:egsr_results}) offers further insights into the model’s behavior. The data suggest that the model performs better for victims exhibiting higher EGS-R scores, a scale associated with pre-clinical PTSD symptoms. Notably, the gap between correctly and incorrectly classified victims widens when speaker information is removed, suggesting that the model is increasingly relying on clinically relevant speech cues--rather than speaker identity--to make predictions. These findings strengthen the hypothesis that pre-clinical trauma manifestations may be encoded in acoustic patterns and can be detected by properly trained models.

A key limitation of the current study is that the participant pool was restricted to victims without a formal PTSD diagnosis, in compliance with ethical requirements to avoid revictimization. As a result, the full spectrum of clinical variability could not be captured. We hypothesize that the presence of more severe trauma (as reflected in higher EGS-R scores) would further improve the ability to detect GBVVC. Future studies should aim to include a broader range of participants to confirm this hypothesis. Another important limitation is the relatively small number of users in the victims' group, which constrains both the generalizability and the statistical power of the model. Additionally, the current approach relies solely on acoustic features extracted from speech, leaving out potentially informative physiological or behavioral modalities--such as heart rate variability, galvanic skin response, linguistics or facial expressions--which could be explored in a multimodal framework to enhance classification robustness and interpretability. From a modeling perspective, the architecture of the DAM blocks was deliberately kept simple to prioritize interpretability and training stability. However, it is possible that more complex architectures--such as transformer-based encoders, hierarchical attention, or recurrent convolutional layers--could better capture fine-grained temporal dynamics and yield improved performance both when reducing the capability of the model to identify speakers and also in the GBVVC detection task. Moreover, the use of fixed-length 1-second frames may discard longer-term dependencies relevant for condition identification; future work could investigate variable-length segments or sequence-to-sequence architectures.

The proposed domain-adversarial scheme also holds potential for broader application beyond GBVVC. It may prove beneficial in other speech-based diagnostic tasks where speaker identity or other personal traits act as confounding variables, particularly in fairness-sensitive contexts such as disease detection across different demographic groups \cite{liEnhancingFairnessDisease2023}.

In summary, this work presents the first application of domain-adversarial learning to detect the GBVVC from speech, extending our prior efforts and contributing a novel approach to this under-investigated field. The proposed model achieves both technical and ethical objectives: it enhances classification performance while mitigating identity-related biases. These findings open promising avenues for the development of speech-based tools that are both effective and privacy-preserving--particularly valuable in sensitive domains such as GBV detection, mental health support, and non-invasive early diagnosis.

\section{Methods}

\subsection{Database}\label{sec:database}
The data used in this study stems from an extended version of the WEMAC database \cite{miranda2024wemac}, a multimodal corpus for affective computing research. The original WEMAC dataset includes speech recordings and biosignals--Blood Volume Pulse (BVP), Galvanic Skin Response (GSR), and Body Temperature (BT)--from Spanish-speaking women with no known history of gender-based violence victimization (non-GBVV). These recordings were captured under controlled laboratory conditions while participants were exposed to 14 emotionally evocative audiovisual stimuli via a Virtual Reality (VR) headset. Participants also provided self-reported emotional annotations following each stimulus.
The original participant pool comprised 100 non-GBVV women aged 20 to 77 years ($mean = 39.92$, $SD = 14.26$), evenly distributed across five predefined age groups: G1 ($18-24$), G2 ($25-34$), G3 ($35-44$), G4 ($45-54$), and G5 ($\geq 55$). Most participants were of Spanish nationality. Following each stimulus, participants responded to two open-ended questions regarding their emotional state, with their answers recorded for speech analysis.
In addition to this publicly available dataset, an ethically approved extension includes speech and questionnaire data from 39 women with a documented history of GBV victimization. For the purposes of this study, participants were considered GBV victims based on self-identification and subsequent confirmation by a certified clinical psychologist specialized in trauma. These data are not publicly released to ensure the participants’ privacy and safety.

All participants completed a biopsychosocial questionnaire that covered sociodemographic information, habits, and lifestyle aspects \cite{wemac_questionnaire}. GBVV participants also completed a second, non-standardized psychological questionnaire designed ad hoc by a licensed psychologist specializing in trauma and GBV. This instrument aimed to gather nuanced information on personal history, relationships, well-being, and trauma. To avoid revictimization, participants displaying symptoms of severe psychological distress were excluded. Specifically, a subset of questions was derived from the standardized EGS-R questionnaire \cite{echeburua2016egsr}, a validated tool for assessing PTSD. Participants scoring above 20 on the EGS-R were excluded from the study.

For classification purposes, we employed a balanced dataset comprising the 39 GBVV participants and an age-matched subset of 39 non-GBVV participants. To mitigate the imbalance in speech segment lengths between groups (GBVV responses tended to be longer), we subsampled the GBVV recordings by randomly removing one out of every four audio segments. All speech signals were resampled to \SI{16}{\kilo\hertz} and normalized using z-score normalization per speaker. The final dataset consisted of 19,128 GBVV samples and 17,400 non-GBVV samples, each 1 second in duration.

Acoustic features were extracted using the Python library \texttt{librosa}--code for feature extraction available in Supplementary Materials Section 5. From each 1-second sample, a total of 19 standard low-level descriptors were computed using a 20 ms window and 10 ms overlap. These include 13 Mel-Frequency Cepstral Coefficients (MFCCs), Root Mean Square (RMS) energy, Zero Crossing Rate, Spectral Centroid, Spectral Roll-off, Spectral Flatness, and Pitch. For each descriptor, both mean and standard deviation were calculated, resulting in a 38-dimensional feature vector per second, following methodologies established in \cite{reyner22iberspeech, reyner23interspeech}. Other feature sets were not considered, as our previous ablation studies \cite{reyner23interspeech} demonstrated that \texttt{librosa}-based features yielded superior performance.

\subsection{Proposed Model Architecture}

The proposed model architecture was implemented in Python using the \texttt{TensorFlow Keras} library--code available in the Supplementary Materials, Section 5. It comprises three main components: an encoder, a speaker identification block (SPK), and a condition classification block (COND). The model was trained under a Leave-One-Subject-Out (LOSO) cross-validation scheme to ensure speaker independence.

\begin{itemize}
    \item Encoder. Projects the 38-dimensional acoustic features into a 128-dimensional embedding space. This component is optimized to extract information relevant for GBVV classification while removing speaker identity traits.
    \item Speaker Identification Block (SPK). Trained to classify the speaker identity. It plays a key role during adversarial training by propagating gradients back through a gradient reversal layer, encouraging the encoder to eliminate speaker-specific information.
    \item Condition Classification Block (COND). Classifies the speech sample as belonging to a GBVV or non-GBVV speaker.
\end{itemize}

A schematic representation of the full architecture, including the interconnection of blocks, is shown in the next section (Section~\ref{sec:advtraining}, Figure \ref{fig:drawio}, and layer-level details are available in Supplementary Materials (Section 6: Figure 11). Hyperparameters used in training are reported in Table \ref{tab:hyperparameters}, and were selected based on a series of ablation studies designed to assess the contribution of different architectural and training choices. These studies, which are discussed in detail in the Supplementary Materials (Section 4), guided the optimization of the model.
\begin{table}[htpb]
\renewcommand{\arraystretch}{1.15}
\centering
\resizebox{.3\columnwidth}{!}{
 \begin{tabular}{c || c} 
 \hline
 Number of Epochs & $100$ \\ 
 Optimizer & Momentum SGD \\
 Batch size & $16$ \\
 Lambda & $0.2$ \\
 Starter learning rate &  $10^{-9}$ \\
 Decay steps & $10000$ \\
 \hline
 \end{tabular}}
\centering
\caption{Training hyperparameters of the proposed models.}
\label{tab:hyperparameters}
\end{table}

\subsection{Domain-Adversarial Training Strategy}
The primary goal of this work is to assess whether it is possible to distinguish victims of gender-based violence (GBVV) from non-victims using speech-based features, without relying on speaker-specific information. To achieve this, we adopt a domain-adversarial training strategy that explicitly aims to preserve information relevant to the GBVV condition while suppressing speaker identity cues. This section details the proposed training approach.

\subsubsection{Initialization Phase: Isolated Models}
Before introducing adversarial training, the encoder, the condition classifier, and the speaker classifier are independently pre-trained. These initial models are referred to as the Isolated Condition Model (ICM) and the Isolated Speaker Model (ISM). Their architectures are illustrated in Figure~\ref{fig:figiso}.
\begin{figure}[htpb]
\centering
\includegraphics[trim={1cm 7.5cm 11cm 0.5cm },clip,width=0.7\textwidth]{FigISO.drawio.pdf}
\caption{Isolated Model Architectures. Black, thinner arrows indicate the forward pass. Thicker, colored arrows denote backpropagation flows.
Top: Isolated Condition Model (ICM). Bottom: Isolated Speaker Model (ISM).}
\label{fig:figiso}
\end{figure}

The purpose of this initialization is twofold. First, it ensures that each module starts from a meaningful set of parameters, thereby facilitating convergence during adversarial training. Second, it establishes baseline models for speaker and condition classification, which are essential to rigorously assess the impact of domain-adversarial training.

\subsubsection{Adversarial Training Procedure}\label{sec:advtraining}
\begin{figure}[htpb]
\centering
\includegraphics[trim={0cm 1cm 7cm 1.2cm},clip,width=0.85\textwidth]{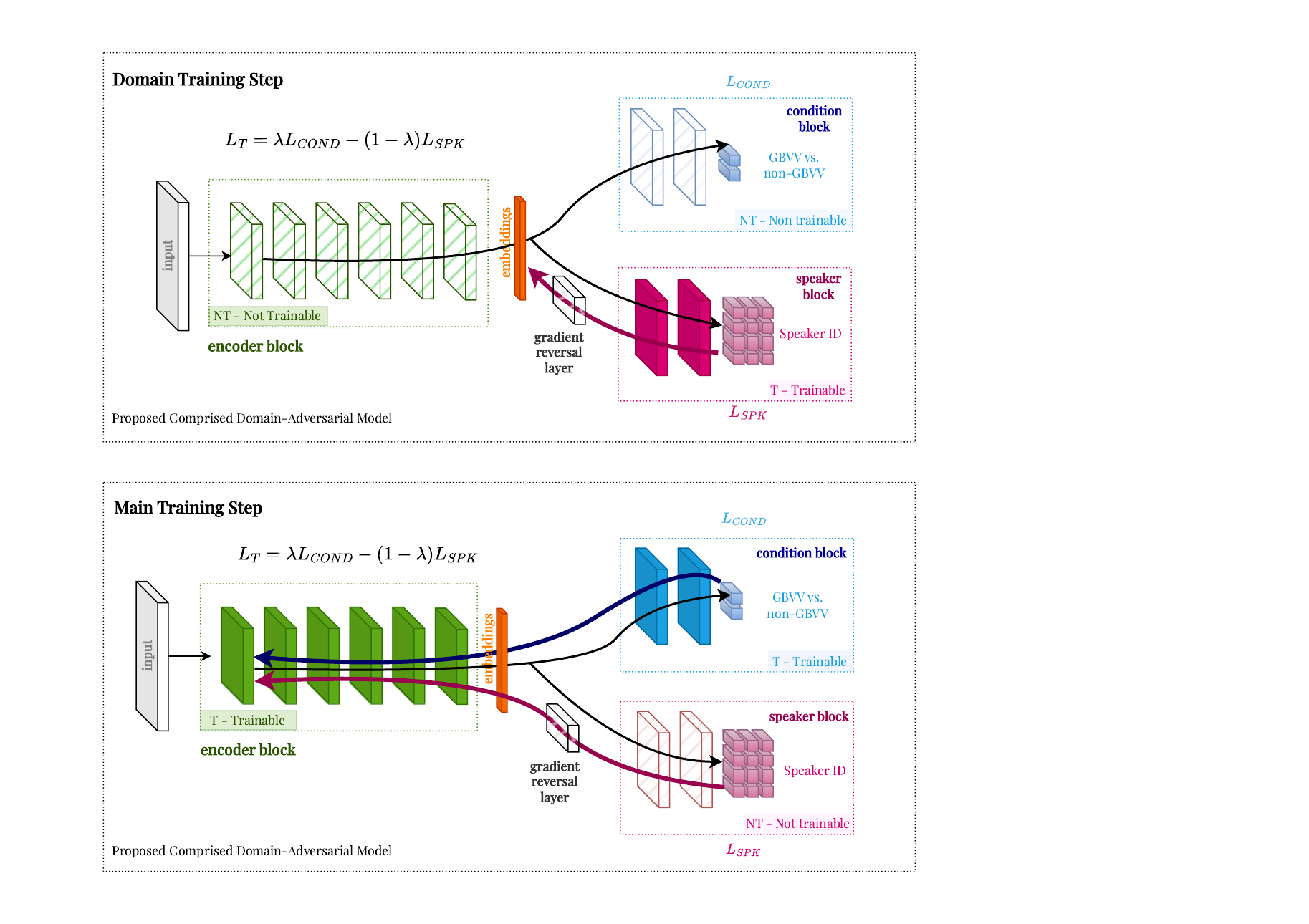}
\caption{Domain-Adversarial Model Architecture. Black arrows indicate the forward pass; colored arrows denote backpropagation. Solid blocks are trainable in the respective step, while dashed blocks are frozen. Top: domain step. Bottom: main step.}
\label{fig:drawio}
\vspace{-5pt}
\end{figure}

Once the ICM and ISM are initialized, we proceed with the domain-adversarial training of the full model (Figure~\ref{fig:drawio}). Each training epoch is composed of three sequential steps: one domain step followed by two main steps.

\emph{1. Domain Step: Speaker Specialization}

The goal of this step is to make the speaker block specialize in speaker identification while ensuring that the encoder does not contribute to this task. To achieve this, the encoder is frozen (non-trainable), and only the speaker classifier is trained using embeddings generated by the fixed encoder. The condition classifier is not used at this stage.

By training the speaker block in isolation, it learns to exploit whatever speaker information remains in the embeddings. This implicitly pushes the adversarial process: in later steps, the encoder will be encouraged to remove this information. The speaker classification loss is computed using categorical cross-entropy over all training speakers (excluding the held-out test speaker), and backpropagation stops at the gradient reversal layer, preventing updates to the encoder.

\emph{2. Main Steps (x2): GBVV Detection and Speaker Information Removal}

The subsequent two steps focus on the main task: detecting GBVV while suppressing speaker identity. Here, the encoder, and condition classifier are all active while the speaker block is frozen. The model is trained to simultaneously predict the GBVV condition while unlearning speaker traits thanks to the gradient reversal layer.

At the core of this mechanism is a custom loss function defined as:
\begin{center}
$L_{T} = \lambda L_{\text{COND}} - (1 - \lambda)L_{\text{SPK}}$
\end{center}
where $L_{\text{COND}}$ is the sparse categorical cross-entropy loss for GBVV condition classification, and $L_{\text{SPK}}$ is the categorical cross-entropy for speaker classification. The hyperparameter $\lambda$ regulates the trade-off between the two objectives; after several ablation studies detailed in the Supplementary Materials (Section 4), the parameter was set to $\lambda = 0.2$ to prioritize removal of speaker information over condition classification.

The gradient reversal layer plays a key role in this setup: it inverts the gradient of the speaker loss before it reaches the encoder. Consequently, the encoder is trained to maximize speaker classification error (i.e., to confuse the speaker classifier), while minimizing the GBVV classification loss. This adversarial mechanism encourages the encoder to produce embeddings that are invariant to speaker identity and discriminative for the GBVV condition.

\emph{Iterative Optimization}

This three-step process is repeated for each of the 100 training epochs. Each epoch comprises one domain step, where only the speaker classifier is trained and the encoder is frozen; followed by two consecutive main steps, where all components are updated adversarially based on the custom loss.
This iterative approach promotes stable training and ensures a gradual removal of speaker information from the learnt embeddings, enabling speaker-independent detection of GBVV.


\section*{Data Availability}
The data comes from a public available source at Miranda Calero, J.A., Gutiérrez-Martín, L., Rituerto-González, E. et al. WEMAC: Women and Emotion Multi-modal Affective Computing dataset. Sci Data 11, 1182 (2024). \url{https://doi.org/10.1038/s41597-024-04002-8} \cite{miranda2024wemac}


\section*{Acknowledgements}

This work has been partially supported by the SAPIENTAE4Bindi Grant PDC2021-121071-I00 funded by MICIU/AEI/10.130
39/501100011033; by the European Union ''NextGenerationEU/PRTR'', PID2021-125780NB-I00 funded by MICIU/AEI/10.1
3039\/501100011033 and, by “ERDF A way of making Europe”; the Federal Ministry of Education and Research of Germany (Bundesministerium für Bildung und Forschung [BMBF]) and the Ministry of Bavaria within the initial phase of the German Center for Mental Health (DZPG), Grant: 01EE2303A.

\section*{Author contributions statement}


\textbf{Emma Reyner-Fuentes} contributed to conceptualization, methodology, software, validation, formal analysis, investigation, data curation, writing—original draft, writing—review and editing, visualization, and supervision.

\textbf{Esther Rituerto-González} contributed to conceptualization, methodology, validation, data curation, writing—original draft, writing—review and editing, and supervision.

\textbf{Carmen Peláez-Moreno} contributed to conceptualization, methodology, investigation, resources, writing—review and editing, supervision, project administration, and funding acquisition.

\section*{Additional information}

\textbf{Accession codes}
\begin{itemize}
    \item WEMAC dataset:
    \url{https://edatos.consorciomadrono.es/dataverse/empatia}
    \item Feature Extraction process:
    \url{https://github.com/BINDI-UC3M/wemac_dataset_signal_processing/tree/master/speech_processing}
    \item Adversarial Training:
    \url{https://github.com/emmareyner/AdversarialTraining}
\end{itemize}

\noindent\textbf{Competing interests.} The authors declare no competing interests.


\begin{thebibliography}{10}
\urlstyle{rm}
\expandafter\ifx\csname url\endcsname\relax
  \def\url#1{\texttt{#1}}\fi
\expandafter\ifx\csname urlprefix\endcsname\relax\def\urlprefix{URL }\fi
\expandafter\ifx\csname doiprefix\endcsname\relax\def\doiprefix{DOI: }\fi
\providecommand{\bibinfo}[2]{#2}
\providecommand{\eprint}[2][]{\url{#2}}

\bibitem{eige2023}
\bibinfo{author}{{European Institute for Gender Equality}}.
\newblock \bibinfo{title}{What is gender-based violence?} (\bibinfo{year}{2023}).

\bibitem{oram2017violence}
\bibinfo{author}{Oram, S.}, \bibinfo{author}{Khalifeh, H.} \& \bibinfo{author}{Howard, L.~M.}
\newblock \bibinfo{journal}{\bibinfo{title}{Violence against women and mental health}}.
\newblock {\emph{{The Lancet Psychiatry}}} \textbf{\bibinfo{volume}{4}}, \bibinfo{pages}{159--170} (\bibinfo{year}{2017}).

\bibitem{escriba2010IPV}
\bibinfo{author}{Escribà-Agüir, V.} \emph{et~al.}
\newblock \bibinfo{journal}{\bibinfo{title}{Partner violence and psychological well-being: Buffer or indirect effect of social support}}.
\newblock {\emph{{Psychosomatic medicine}}} \textbf{\bibinfo{volume}{72}}, \bibinfo{pages}{383--9}, \doiprefix\url{10.1097/PSY.0b013e3181d2f0dd} (\bibinfo{year}{2010}).

\bibitem{ferrari2014DVA}
\bibinfo{author}{Ferrari, G.} \emph{et~al.}
\newblock \bibinfo{journal}{\bibinfo{title}{Domestic violence and mental health: A cross-sectional survey of women seeking help from domestic violence support services}}.
\newblock {\emph{{Global health action}}} \textbf{\bibinfo{volume}{7}}, \bibinfo{pages}{25519}, \doiprefix\url{10.3402/gha.v7.25519} (\bibinfo{year}{2014}).

\bibitem{spencerHealthEffectsAssociated2023}
\bibinfo{author}{Spencer, C.~N.} \emph{et~al.}
\newblock \bibinfo{journal}{\bibinfo{title}{Health effects associated with exposure to intimate partner violence against women and childhood sexual abuse: A {{Burden}} of {{Proof}} study}}.
\newblock {\emph{{Nature Medicine}}} \textbf{\bibinfo{volume}{29}}, \bibinfo{pages}{3243--3258}, \doiprefix\url{10.1038/s41591-023-02629-5} (\bibinfo{year}{2023}).

\bibitem{haeringDisentanglingSexDifferences2024}
\bibinfo{author}{Haering, S.} \emph{et~al.}
\newblock \bibinfo{journal}{\bibinfo{title}{Disentangling sex differences in {{PTSD}} risk factors}}.
\newblock {\emph{{Nature Mental Health}}} \textbf{\bibinfo{volume}{2}}, \bibinfo{pages}{605--615}, \doiprefix\url{10.1038/s44220-024-00236-y} (\bibinfo{year}{2024}).

\bibitem{chandan2019ipv}
\bibinfo{author}{Chandan, J.} \emph{et~al.}
\newblock \bibinfo{journal}{\bibinfo{title}{Female survivors of intimate partner violence and risk of depression, anxiety and serious mental illness}}.
\newblock {\emph{{The British journal of psychiatry : the journal of mental science}}} \textbf{\bibinfo{volume}{217}}, \bibinfo{pages}{1--6}, \doiprefix\url{10.1192/bjp.2019.124} (\bibinfo{year}{2019}).

\bibitem{shen2019intimate}
\bibinfo{author}{Shen, S.} \& \bibinfo{author}{Kusunoki, Y.}
\newblock \bibinfo{journal}{\bibinfo{title}{Intimate partner violence and psychological distress among emerging adult women: A bidirectional relationship}}.
\newblock {\emph{{Journal of Women's Health}}} \textbf{\bibinfo{volume}{28}}, \bibinfo{pages}{1060--1067} (\bibinfo{year}{2019}).

\bibitem{who2021}
\bibinfo{author}{{World Health Organization}}.
\newblock \bibinfo{title}{Violence against women} (\bibinfo{year}{2021}).

\bibitem{undoc1993}
\bibinfo{author}{{United Nations}}.
\newblock \bibinfo{title}{Declaration on the elimination of violence against women} (\bibinfo{year}{1993}).

\bibitem{leyorganica2004}
\bibinfo{title}{Artículo 1 de la ley orgánica 1/2004, de 28 de diciembre, de medidas de protección integral contra la violencia de género}.

\bibitem{echeburua2016egsr}
\bibinfo{author}{Echeburúa, E.} \emph{et~al.}
\newblock \bibinfo{journal}{\bibinfo{title}{Escala de gravedad de síntomas revisada (egs-r) del trastorno de estrés postraumático según el dsm-5: propiedades psicométricas}}.
\newblock {\emph{{Terapia Psicológica}}} \textbf{\bibinfo{volume}{34}}, \bibinfo{pages}{111--128} (\bibinfo{year}{2016}).

\bibitem{shanmugamQuantifyingDisparitiesIntimate2024}
\bibinfo{author}{Shanmugam, D.}, \bibinfo{author}{Hou, K.} \& \bibinfo{author}{Pierson, E.}
\newblock \bibinfo{journal}{\bibinfo{title}{Quantifying disparities in intimate partner violence: A machine learning method to correct for underreporting}}.
\newblock {\emph{{npj Women's Health}}} \textbf{\bibinfo{volume}{2}}, \bibinfo{pages}{1--13}, \doiprefix\url{10.1038/s44294-024-00011-5} (\bibinfo{year}{2024}).

\bibitem{chenUsingMachineLearning2023}
\bibinfo{author}{Chen, Z.} \emph{et~al.}
\newblock \bibinfo{journal}{\bibinfo{title}{Using machine learning to estimate the incidence rate of intimate partner violence}}.
\newblock {\emph{{Scientific Reports}}} \textbf{\bibinfo{volume}{13}}, \bibinfo{pages}{5533}, \doiprefix\url{10.1038/s41598-023-31846-8} (\bibinfo{year}{2023}).

\bibitem{koops2023speech}
\bibinfo{author}{Koops, S.} \emph{et~al.}
\newblock \bibinfo{journal}{\bibinfo{title}{Speech as a biomarker for depression}}.
\newblock {\emph{{CNS \& Neurological Disorders-Drug Targets (Formerly Current Drug Targets-CNS \& Neurological Disorders)}}} \textbf{\bibinfo{volume}{22}}, \bibinfo{pages}{152--160} (\bibinfo{year}{2023}).

\bibitem{scherer2013suicidal}
\bibinfo{author}{Scherer, S.}, \bibinfo{author}{Pestian, J.} \& \bibinfo{author}{Morency, L.-P.}
\newblock \bibinfo{title}{Investigating the speech characteristics of suicidal adolescents}.
\newblock In \emph{\bibinfo{booktitle}{2013 IEEE International Conference on Acoustics, Speech and Signal Processing}}, \bibinfo{pages}{709--713}, \doiprefix\url{10.1109/ICASSP.2013.6637740} (\bibinfo{year}{2013}).

\bibitem{belouali2021acoustic}
\bibinfo{author}{Belouali, A.} \emph{et~al.}
\newblock \bibinfo{journal}{\bibinfo{title}{Acoustic and language analysis of speech for suicidal ideation among us veterans}}.
\newblock {\emph{{BioData mining}}} \textbf{\bibinfo{volume}{14}}, \bibinfo{pages}{1--17} (\bibinfo{year}{2021}).

\bibitem{latif2020speech}
\bibinfo{author}{Latif, S.}, \bibinfo{author}{Qadir, J.}, \bibinfo{author}{Qayyum, A.}, \bibinfo{author}{Usama, M.} \& \bibinfo{author}{Younis, S.}
\newblock \bibinfo{journal}{\bibinfo{title}{Speech technology for healthcare: Opportunities, challenges, and state of the art}}.
\newblock {\emph{{IEEE Reviews in Biomedical Engineering}}} \textbf{\bibinfo{volume}{PP}}, \bibinfo{pages}{1--1}, \doiprefix\url{10.1109/RBME.2020.3006860} (\bibinfo{year}{2020}).

\bibitem{wanderley2022detection}
\bibinfo{author}{Wanderley~Espinola, C.}, \bibinfo{author}{Gomes, J.~C.}, \bibinfo{author}{M{\^o}nica Silva~Pereira, J.} \& \bibinfo{author}{dos Santos, W.~P.}
\newblock \bibinfo{journal}{\bibinfo{title}{Detection of major depressive disorder, bipolar disorder, schizophrenia and generalized anxiety disorder using vocal acoustic analysis and machine learning: an exploratory study}}.
\newblock {\emph{{Research on Biomedical Engineering}}} \textbf{\bibinfo{volume}{38}}, \bibinfo{pages}{813--829} (\bibinfo{year}{2022}).

\bibitem{schuller2013paralinguistics}
\bibinfo{author}{Schuller, B.} \emph{et~al.}
\newblock \bibinfo{journal}{\bibinfo{title}{Paralinguistics in speech and language—state-of-the-art and the challenge}}.
\newblock {\emph{{Computer Speech \& Language}}} \textbf{\bibinfo{volume}{27}}, \bibinfo{pages}{4--39} (\bibinfo{year}{2013}).

\bibitem{mura2020ser}
\bibinfo{author}{La~Mura, M.} \& \bibinfo{author}{Lamberti, P.}
\newblock \bibinfo{title}{Human-machine interaction personalization: a review on gender and emotion recognition through speech analysis}.
\newblock In \emph{\bibinfo{booktitle}{2020 IEEE International Workshop on Metrology for Industry 4.0 \& IoT}}, \bibinfo{pages}{319--323}, \doiprefix\url{10.1109/MetroInd4.0IoT48571.2020.9138203} (\bibinfo{year}{2020}).

\bibitem{reyner22iberspeech}
\bibinfo{author}{{Reyner Fuentes}, E.}, \bibinfo{author}{{Rituerto González}, E.}, \bibinfo{author}{Mingueza, C.~L.}, \bibinfo{author}{{Peláez Moreno}, C.} \& \bibinfo{author}{{López Ongil}, C.}
\newblock \bibinfo{title}{{Detecting Gender-based Violence aftereffects from Emotional Speech Paralinguistic Features }}.
\newblock In \emph{\bibinfo{booktitle}{Proc. IberSPEECH 2022}}, \bibinfo{pages}{96--100}, \doiprefix\url{10.21437/IberSPEECH.2022-20} (\bibinfo{year}{2022}).

\bibitem{reyner23interspeech}
\bibinfo{author}{Reyner-Fuentes, E.}, \bibinfo{author}{Rituerto-González, E.}, \bibinfo{author}{Trancoso, I.} \& \bibinfo{author}{Peláez-Moreno, C.}
\newblock \bibinfo{title}{{Prediction of the Gender-based Violence Victim Condition using Speech: What do Machine Learning Models rely on?}}
\newblock In \emph{\bibinfo{booktitle}{Proc. INTERSPEECH 2023}}, \bibinfo{pages}{1768--1772}, \doiprefix\url{10.21437/Interspeech.2023-1771} (\bibinfo{year}{2023}).

\bibitem{ganin2016domainadversarial}
\bibinfo{author}{Ganin, Y.} \emph{et~al.}
\newblock \bibinfo{title}{Domain-adversarial training of neural networks} (\bibinfo{year}{2016}).
\newblock \eprint{1505.07818}.

\bibitem{miranda2024wemac}
\bibinfo{author}{Miranda~Calero, J.~A.} \emph{et~al.}
\newblock \bibinfo{journal}{\bibinfo{title}{{{WEMAC}}: {{Women}} and {{Emotion Multi-modal Affective Computing}} dataset}}.
\newblock {\emph{{Scientific Data}}} \textbf{\bibinfo{volume}{11}}, \bibinfo{pages}{1182}, \doiprefix\url{10.1038/s41597-024-04002-8} (\bibinfo{year}{2024}).

\bibitem{math9080807}
\bibinfo{author}{Castorena, C.~M.} \emph{et~al.}
\newblock \bibinfo{journal}{\bibinfo{title}{Deep neural network for gender-based violence detection on twitter messages}}.
\newblock {\emph{{Mathematics}}} \textbf{\bibinfo{volume}{9}}, \doiprefix\url{10.3390/math9080807} (\bibinfo{year}{2021}).

\bibitem{8685083}
\bibinfo{author}{Subramani, S.} \emph{et~al.}
\newblock \bibinfo{journal}{\bibinfo{title}{Deep learning for multi-class identification from domestic violence online posts}}.
\newblock {\emph{{IEEE Access}}} \textbf{\bibinfo{volume}{7}}, \bibinfo{pages}{46210--46224}, \doiprefix\url{10.1109/ACCESS.2019.2908827} (\bibinfo{year}{2019}).

\bibitem{yallico2022gbv}
\bibinfo{author}{Yallico~Arias, T.} \& \bibinfo{author}{Fabian, J.}
\newblock \bibinfo{title}{Automatic detection of levels of intimate partner violence against women with natural language processing using machine learning and deep learning techniques}.
\newblock In \bibinfo{editor}{Lossio-Ventura, J.~A.} \emph{et~al.} (eds.) \emph{\bibinfo{booktitle}{Information Management and Big Data}}, \bibinfo{pages}{189--205} (\bibinfo{publisher}{Springer International Publishing}, \bibinfo{address}{Cham}, \bibinfo{year}{2022}).

\bibitem{berishaAreReportedAccuracies2022}
\bibinfo{author}{Berisha, V.}, \bibinfo{author}{Krantsevich, C.}, \bibinfo{author}{Stegmann, G.}, \bibinfo{author}{Hahn, S.} \& \bibinfo{author}{Liss, J.}
\newblock \bibinfo{title}{Are reported accuracies in the clinical speech machine learning literature overoptimistic?}
\newblock In \emph{\bibinfo{booktitle}{Interspeech 2022}}, \bibinfo{pages}{2453--2457}, \doiprefix\url{10.21437/Interspeech.2022-691} (\bibinfo{publisher}{{ISCA}}, \bibinfo{year}{2022}).

\bibitem{rutowskiCorpusSizeRequirements2022}
\bibinfo{author}{Rutowski, T.} \emph{et~al.}
\newblock \bibinfo{title}{Toward {{Corpus Size Requirements}} for {{Training}} and {{Evaluating Depression Risk Models Using Spoken Language}}}.
\newblock In \emph{\bibinfo{booktitle}{Interspeech 2022}}, \bibinfo{pages}{3343--3347}, \doiprefix\url{10.21437/Interspeech.2022-10888} (\bibinfo{publisher}{{ISCA}}, \bibinfo{year}{2022}).

\bibitem{millingSpeechNewBlood2022}
\bibinfo{author}{Milling, M.}, \bibinfo{author}{Pokorny, F.~B.}, \bibinfo{author}{{Bartl-Pokorny}, K.~D.} \& \bibinfo{author}{Schuller, B.~W.}
\newblock \bibinfo{journal}{\bibinfo{title}{Is {{Speech}} the {{New Blood}}? {{Recent Progress}} in {{AI-Based Disease Detection From Audio}} in a {{Nutshell}}}}.
\newblock {\emph{{Frontiers in Digital Health}}} \textbf{\bibinfo{volume}{4}} (\bibinfo{year}{2022}).

\bibitem{xuMachineUnlearningSurvey2023}
\bibinfo{author}{Xu, H.}, \bibinfo{author}{Zhu, T.}, \bibinfo{author}{Zhang, L.}, \bibinfo{author}{Zhou, W.} \& \bibinfo{author}{Yu, P.~S.}
\newblock \bibinfo{journal}{\bibinfo{title}{Machine {{Unlearning}}: {{A Survey}}}}.
\newblock {\emph{{ACM Computing Surveys}}} \textbf{\bibinfo{volume}{56}}, \bibinfo{pages}{9:1--9:36}, \doiprefix\url{10.1145/3603620} (\bibinfo{year}{2023}).

\bibitem{li2020adversarial}
\bibinfo{author}{Li, H.}, \bibinfo{author}{Tu, M.}, \bibinfo{author}{Huang, J.}, \bibinfo{author}{Narayanan, S.} \& \bibinfo{author}{Georgiou, P.}
\newblock \bibinfo{title}{Speaker-invariant affective representation learning via adversarial training}.
\newblock In \emph{\bibinfo{booktitle}{ICASSP 2020 - 2020 IEEE International Conference on Acoustics, Speech and Signal Processing (ICASSP)}}, \bibinfo{pages}{7144--7148}, \doiprefix\url{10.1109/ICASSP40776.2020.9054580} (\bibinfo{year}{2020}).

\bibitem{han2018adversarial}
\bibinfo{author}{Han, J.}, \bibinfo{author}{Zhang, Z.}, \bibinfo{author}{Cummins, N.} \& \bibinfo{author}{Schuller, B.}
\newblock \bibinfo{title}{Adversarial training in affective computing and sentiment analysis: Recent advances and perspectives} (\bibinfo{year}{2018}).
\newblock \eprint{1809.08927}.

\bibitem{liEnhancingFairnessDisease2023}
\bibinfo{author}{Li, B.} \emph{et~al.}
\newblock \bibinfo{title}{Enhancing {{Fairness}} in {{Disease Prediction}} by {{Optimizing Multiple Domain Adversarial Networks}}}, \doiprefix\url{10.1101/2023.08.04.551906} (\bibinfo{year}{2023}).

\bibitem{wemac_questionnaire}
\bibinfo{author}{Miranda~Calero, J.~A.} \emph{et~al.}
\newblock \bibinfo{title}{{UC3M4Safety Database - WEMAC: Biopsychosocial questionnaire and informed consent}}, \doiprefix\url{10.21950/U5DXJR} (\bibinfo{year}{2022}).

\end{thebibliography}
\end{document}